\pdfoutput=1
\documentclass[11pt]{article}
\usepackage[svgnames]{xcolor}

\usepackage{acl}

\usepackage{times}
\usepackage{latexsym}
\usepackage{booktabs}
\usepackage{amsmath}
\usepackage{amssymb}
\usepackage{bbm}
\usepackage{multirow}
\usepackage{graphicx}
\usepackage{CJKutf8}

\usepackage[T1]{fontenc}
\usepackage[utf8]{inputenc}

\usepackage{microtype}

\title{Building Chinese Biomedical Language Models via Multi-Level \\Text Discrimination}

\author{Quan Wang$^{1{\color{NavyBlue}{\dagger}}}$\thanks{\hspace{0.15cm}Correspondence to {\tt quanwang1012@gmail.com}.} , Songtai Dai$^{1}$\thanks{\hspace{0.15cm}Contributed equally to this work.} , Benfeng Xu$^{2{\color{NavyBlue}{\dagger}}}$\thanks{\hspace{0.15cm}Work done during internship at Baidu Inc.} , Yajuan Lyu$^1$ \\
  {\bf Yong Zhu$^1$, Hua Wu$^1$, Haifeng Wang$^1$} \\
  $^1$Baidu Inc., Beijing, China \\
  $^2$University of Science and Technology of China, Hefei, China}

\begin{document}
\begin{CJK}{UTF8}{gkai}
\maketitle
\begin{abstract}
Pre-trained language models (PLMs), such as BERT and GPT, have revolutionized the field of NLP, not only in the general domain but also in the biomedical domain. Most prior efforts in building biomedical PLMs have resorted simply to domain adaptation and focused mainly on English. In this work we introduce eHealth, a Chinese biomedical PLM built from scratch with a new pre-training framework. This new framework pre-trains eHealth as a discriminator through both token- and sequence-level discrimination. The former is to detect input tokens corrupted by a generator and recover their original identities from plausible candidates, while the latter is to further distinguish corruptions of a same original sequence from those of others. As such, eHealth can learn language semantics at both token and sequence levels. Extensive experiments on 11 Chinese biomedical language understanding tasks of various forms verify the effectiveness and superiority of our approach. We release the pre-trained model to the public,\footnote{\url{https://github.com/PaddlePaddle/Research/tree/master/KG/eHealth}} and will also release the code later.
%\footnote{\url{https://github.com/PaddlePaddle/Research/tree/master/KG/eHealth}} 
\end{abstract}

\section{Introduction}
Pre-trained language models (PLMs) such as BERT \cite{devlin2019:bert} and its variants \cite{yang2019:xlnet,liu2019:roberta} have revolutionized the field of NLP, establishing new state-of-the-art on conventional language understanding and generation tasks. Following the great success in the general domain, researchers have started to investigate building domain-specific PLMs in highly specialized domains, {\it e.g.}, science \cite{beltagy2019:scibert}, law \cite{chalkidis2020:legal-bert}, or finance \cite{liu2020:finbert}. Biomedicine and healthcare, as a field with large, rapidly growing volume of free text and continually increasing demand for text mining, has received massive attention and achieved rapid progress.

Biomedical PLMs are typically built by adapting a general-domain PLM to the biomedical domain with (almost) the same model architecture and training objectives, as exemplified by BioBERT \cite{lee2020:biobert}, PubMedBERT \cite{gu2020:pubmedbert}, and BioELECTRA \cite{kanakarajan2021:bioelectra}. This domain adaptation is achieved via either {\it continual pre-training} on in-domain text \cite{gururangandon2020:biomedroberta}, or {\it pre-training from scratch} further with an in-domain vocabulary \cite{gu2020:pubmedbert,lewis2020:biolm}, which has shown to be particularly useful for English biomedical text understanding. 

As for the Chinese biomedical field, MC-BERT \cite{zhang2020:mcbert} and PCL-MedBERT are two initial attempts that continually pre-train a general-domain BERT on in-domain text. But unfortunately they fail to achieve satisfactory performance compared with their general-domain rivals \cite{zhang2021:cblue}. SMedBERT \cite{zhang2021:smedbert} and EMBERT \cite{cai2021:embert} also continually pre-train from the general-domain BERT, but in knowledge-enhanced fashions. These two models rely on external (and often private) knowledge and have not been released to the public yet. So far there is still a lack of publicly available, high-quality biomedical PLMs in Chinese. 

In this paper we present {\bf eHealth}, a Chinese language representation model pre-trained over large-scale biomedical text corpora. Unlike most previous studies that simply resort to direct domain adaptation, we build eHealth with a new self-supervised learning framework, which, similar to ELECTRA \cite{clark2020:electra}, consists of a discriminator and a generator. The generator is to produce corrupted input, and the discriminator, as the final target encoder, is trained via multi-level text discrimination. Specifically, we employ (i) {\it token-level discrimination} that discriminates corrupted tokens from original ones, and (ii) {\it sequence-level discrimination} that further discriminates corruptions of a same original sequence from those of others in a contrastive learning fashion \cite{chen2020:simclr}. This multi-level discrimination enables eHealth to learn language semantics at both token and sequence levels. 

As a new Chinese biomedical PLM, eHealth has two distinguishing features: built-from-scratch and easy-to-deploy. By the former we mean that unlike all prior arts that start pre-training from a general-domain Chinese BERT and directly use the associated vocabulary, eHealth is pre-trained entirely from scratch with a newly built in-domain vocabulary. This vocabulary, as we will show later in our experiments, can better tokenize biomedical text and may lead to better understanding of such text. And by the latter we mean that eHealth relies solely on the text itself, requiring no additional retrieval, linking, or encoding of relevant knowledge as those knowledge-enhanced models do, and thereby could be applied rather easily during fine-tuning.

We evaluate eHealth on 11 diversified Chinese biomedical language understanding tasks, including (i) the 8 tasks of text classification and matching, medical information extraction, and medical term normalization from the CBLUE benchmark \cite{zhang2021:cblue}, and (ii) another 3 medical question answering tasks cMedQNLI \cite{zhang2020:mcbert}, webMedQA \cite{he2019:webmedqa}, and NLPEC \cite{li2020:nlpec}. Experimental results reveal that eHealth, as a standard base-sized model pre-trained from scratch on biomedical corpora, consistently outperforms previous state-of-the-art PLMs in almost all cases, no matter those from the general domain or biomedical domain, and no matter those base-sized or even large-sized.

The main contributions of this work are two-fold. Firstly, we propose a new Chinese biomedical PLM and release the pre-trained model to the public. This new model shows superior ability in Chinese biomedical text understanding and is easy to deploy. Secondly, we devise a new algorithm for language model pre-training and verify its effectiveness in the biomedical domain. This pre-training algorithm is quite generic and may be readily adapted to other domains beyond biomedicine. We leave such exploration open to future work.

\section{Background}
Before diving into the details of our approach, we briefly discuss related studies on building PLMs in general and biomedical domains.

\paragraph{General Domain PLMs.}
Recent years have seen remarkable success of PLMs in the field of NLP. These PLMs are typically built with self-supervised learning over massive unlabeled text in the general domain, {\it e.g.}, Wikipedia, newswire, or Web articles \cite{radford2018:gpt}. {\it Masked language modeling} (MLM), which trains a model to recover the identities of a small subset of masked-out tokens (typically 15\%), is the most prevailing self-supervised objective, first introduced in BERT \cite{devlin2019:bert} and then widely adopted by follow-up studies \cite{liu2019:roberta,lan2020:albert,joshi2020:spanbert,sun2020:ernie20}. Despite their effectiveness and popularity, MLM-based approaches can only learn from those 15\% masked-out tokens per input, and therefore incur high compute costs.

To address this low efficiency issue, ELECTRA \cite{clark2020:electra} uses a new pre-training framework. Specifically, it corrupts an input sequence by replacing some of the tokens with plausible alternatives sampled from an auxiliary generator, and trains a discriminator to predict for each token in that sequence whether it is original or replaced, {\it i.e.}, {\it replaced token detection} (RTD). As the discriminator can learn from all input tokens rather than just 15\% of them, ELECTRA enjoys better efficiency and accelerates training.

While achieving empirical success, there are concerns about whether the over-simplified RTD task of ELECTRA, as a binary classification problem, is informative enough for language modeling \cite{aroca2020:losses}. \citet{xu2020:metacontroller} and \citet{shen2021:teams} thus proposed training the model via a generalization of RTD while a simplification of MLM, by recovering for each token its original identity from a few plausible candidates, rather than from the whole vocabulary.

%\citet{xu2020:metacontroller} thus proposed to replace RTD with a {\it multi-token selection} (MTS) task, which selects for each token its original identity from a few plausible candidates, rather than from the whole vocabulary. This new task could be viewed as a generalization of RTD while a simplification of MLM. \citet{shen2021:teams} later investigated training MTS together with RTD in a multi-task fashion.

Another limitation of ELECTRA is that it is pre-trained solely at the token level but lacks semantics at the sequence level. Incorporating sequence level signals, {\it e.g.}, next sentence prediction \cite{devlin2019:bert}, sentence order prediction \cite{lan2020:albert}, and sentence contrastive learning \cite{fang2020:cert,meng2021:coco}, has been widely accepted in the community and shown to be beneficial in specific tasks \cite{lewis2020:bart,guu2020:realm}.

In this paper, to build a Chinese biomedical PLM, we employ the ELECTRA framework which favors the efficiency of pre-training. Within this framework, we strengthen the oversimplified RTD task and introduce sequence-level signals, which further improves the quality of pre-training.

\paragraph{Biomedical PLMs.}
Continual pre-training is perhaps the most straightforward way to build biomedical PLMs, in which the model weights are initialized from a well-trained general-domain model and the same vocabulary is used \cite{alsentzer2019:clinicalbert,lee2020:biobert}. Also, there are findings showing that pre-training from scratch using domain specific data along with domain specific vocabulary  would bring further improvements, particularly in English \cite{gu2020:pubmedbert,lewis2020:biolm}. Early attempts focused on adapting BERT, while recent studies have switched to its modern variants like RoBERTa, ALBERT, and ELECTRA \cite{kanakarajan2021:bioelectra,alrowili2021:biom}. 

While great efforts have been made to build English biomedical PLMs, there is only a few studies discussing building biomedical PLMs in Chinese, {\it e.g.}, MC-BERT \cite{zhang2020:mcbert}, SMedBERT \cite{zhang2021:smedbert}, and EMBERT \cite{cai2021:embert}, all resumed from a general-domain BERT, with the latter two further in knowledge-enhanced fashions.\footnote{Actually there are two versions of EMBERT, one initialized with BERT and the other with MC-BERT, which is also resumed from BERT.} Models like this typically require extra knowledge and consequently the retrieval, linking, and encoding of such knowledge. They are not that easy to be applied to downstream tasks.

\section{Methodology}
\begin{figure}[t]
\centering
\includegraphics[width=0.48\textwidth]{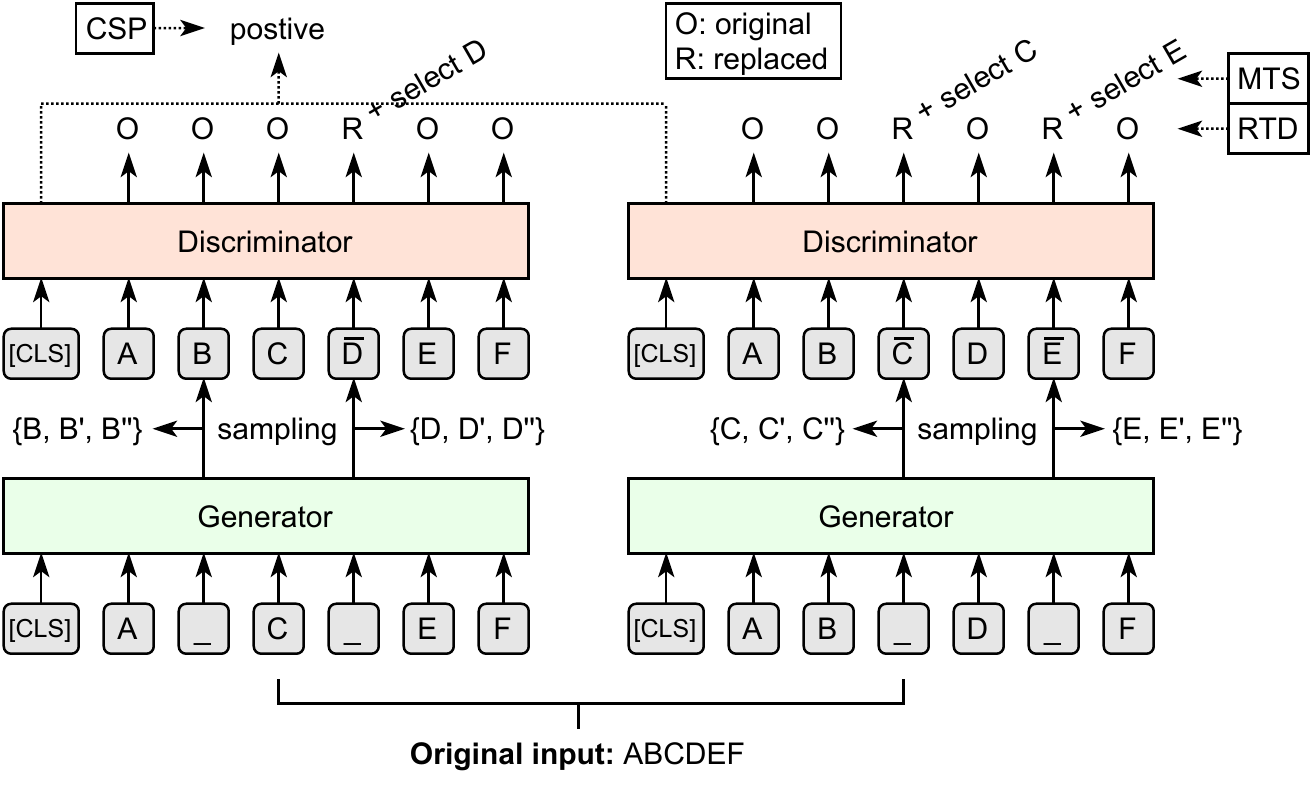}
\caption{Overview of eHealth. Each input sequence is corrupted twice independently by the generator. These two corruptions are fed into the discriminator for replaced token detection (RTD) and multi-token selection (MTS), {\it i.e.}, token-level discrimination. And they also form a positive pair for contrastive sequence prediction (CSP), {\it i.e.}, sequence-level discrimination.}
\label{fig:overview}
\end{figure}

This section presents eHealth, a Chinese language model pre-trained from biomedical text. It in general follows the generator-discriminator framework of ELECTRA, where the generator $G$ is introduced to construct pre-training signals and the discriminator $D$ is used as the final target encoder. But unlike ELECTRA that merely adopts a token-level binary classification to train the discriminator, we train it with (i) a more informative token-level discrimination, and (ii) another sequence-level discrimination. The overview of eHealth is illustrated in Figure~\ref{fig:overview}. 

\subsection{Generator}\label{subsec:generator}
The generator $G$ is a Transformer encoder \cite{vaswani2017:transformer} trained by {\it masked language modeling} (MLM). Given an input sequence $\mathbf{x}\!=\![x_1, \cdots, x_n]$, it first selects a random set of positions to mask out and replaces tokens at these positions with a special symbol {\tt [MASK]}.\footnote{Typically 15\% of the tokens are masked out, among which 80\% are replaced with {\tt [MASK]}, 10\% replaced with a random token, and 10\% kept unchanged.} This masked sequence, denoted as $\mathbf{x}^{M}$, is then passed into the Transformer encoder to produce contextualized representations $h_G(\mathbf{x}^{M})$, and thereafter a softmax layer to predict the original identities of those masked-out tokens: 
\begin{equation}\label{eq:mlm-prob}
\small
p_G(x_t | \mathbf{x}^{M}) = \frac{\exp\big(e(x_t)^{\rm T} h_G(\mathbf{x}^{M})_t\big)}{\sum_{x' \in V} \exp\big(e(x')^{\rm T} h_G(\mathbf{x}^{M})_t\big)}.
\end{equation}
Here, $p_G(x_t | \mathbf{x}^{M})$ is the probability that $G$ predicts token $x_t$ appears at the $t$-th masked position in $\mathbf{x}^M$, $h_G(\mathbf{x}^{M})_t$ the contextualized representation for that position, $e(\cdot)$ the embedding lookup operation on each token, and $V$ the vocabulary of all tokens. The corresponding loss function is:
\begin{equation}\label{eq:mlm-loss}
\small
\mathcal{L}_{\rm MLM} (\mathbf{x}, \mathbf{x}^M; G) = \sum_{t:x_t^M={\tt [MASK]}} \!\!\!\! - \log p_G(x_t | \mathbf{x}^{M}),
\end{equation}
where the summation is taken only over the masked positions. The generator is used to construct pre-training signals for the discriminator, and will be discarded after pre-training. 

\subsection{Discriminator}\label{subsec:discriminator}
The discriminator $D$, as our final target encoder, is also a Transformer architecture. It takes as input corrupted sequences constructed by the generator, and is trained through two-level text discrimination, {\it i.e.}, token-level and sequence-level, so as to encode language semantics at both levels.

\paragraph{Token-Level Discrimination.}
We consider two token-level tasks: {\it replaced token detection} (RTD) and {\it multi-token selection} (MTS). RTD is the standard pre-training task of ELECTRA, which detects replaced tokens in a corrupted sequence, and MTS further selects original identities for those replaced tokens. Specifically, given input sequence $\mathbf{x}$ and its masked version $\mathbf{x}^M$, for each masked position $t$, we sample a token from the generator's prediction $\hat{x}_t \sim p_G(x_t | \mathbf{x}^{M})$ (cf. Eq.~(\ref{eq:mlm-prob})), replace the original token $x_t$ with $\hat{x}_t$, and create a corrupted sequence $\mathbf{x}^R$. We also create a set of candidate tokens, denoted as $S_t$, for each masked position $t$, by drawing $k$ non-original tokens from $p_G(x_t | \mathbf{x}^{M})$ along with the original token $x_t$. The discriminator $D$ encodes the corrupted sequence $\mathbf{x}^R$ and produces contextualized representations $h_D(\mathbf{x}^{R})$.

RTD learns to discriminate whether each token in $\mathbf{x}^R$ is original or replaced, {\it i.e.}, coming from the true data distribution or the generator distribution. It uses a sigmoid layer on top of $h_D(\mathbf{x}^{R})$ to perform this binary classification, where the probability that $x_t^R$ matches the original token $x_t$ is determined as:
\begin{equation}\label{eq:rtd-prob}
\small
p_D(x_t^R = x_t) = \frac{1}{1+ \exp ( - \mathbf{w}^{\rm T} h_D(\mathbf{x}^{R})_t )},
\end{equation}
and the corresponding loss function is:
\begin{small}
\begin{eqnarray}\label{eq:rtd-loss}
\mathcal{L}_{\rm RTD} (\mathbf{x}, \mathbf{x}^R;\! D)  \!=\! \sum_{t=1}^n \big[ - \!\! \mathbbm{1}(x_t^R \!=\! x_t) \log p_D(x_t^R \!=\! x_t) \notag \\
- \mathbbm{1}(x_t^R \!\neq\! x_t) \log (1- p_D(x_t^R \!=\! x_t)) \big].
\end{eqnarray}
\end{small}As merely a binary classification task, RTD might not be informative enough for language modeling. 
%\cite{aroca2020:losses,xu2020:metacontroller,shen2021:teams}

MTS strengthens RTD by training the discriminator to further recover original identities of those replaced tokens. For each position $t$ where the token is replaced, {\it i.e.}, $x_t^R \neq x_t$, MTS corrects the token and recovers its original identity from candidate set $S_t$. The probability of picking the original identity $x_t$ out of $S_t$ for the correction is:
\begin{equation}\label{eq:mts-prob}
\small
p_D(x_t | \mathbf{x}^{R}, S_t) = \frac{\exp\big(e(x_t)^{\rm T} h_D(\mathbf{x}^{R})_t\big)}{\sum_{x' \in S_t} \exp\big(e(x')^{\rm T} h_D(\mathbf{x}^{R})_t\big)},
\end{equation}
where $e(\cdot)$ is again the embedding lookup operation. The loss function is defined as: 
\begin{equation}\label{eq:mts-loss}
\small
\mathcal{L}_{\rm MTS} (\mathbf{x}, \mathbf{x}^R, \mathcal{S}; D) = \sum_{t:x_t^R \neq x_t} \!\!\! - \log p_D(x_t | \mathbf{x}^{R}, S_t),
\end{equation}
where $\mathcal{S}=\{S_t\}_{t:x_t^R \neq x_t}$ is a collection of candidate sets at all positions with replaced tokens, and the summation is taken only over these positions. MTS is essentially a $(k+1)$-class classification problem. It is more challenging than RTD and hence pushes the discriminator to learn representations that encode richer semantic information \cite{xu2020:metacontroller,shen2021:teams}. 

\paragraph{Sequence-Level Discrimination.}
$\!\!$Besides token-level tasks, we consider a sequence-level task in addition, {\it i.e.}, {\it contrastive sequence prediction} (CSP) which learns to discriminate corruptions of a single original sequence from those of the others. CSP employs a classic contrastive learning framework \cite{chen2020:simclr}. Specifically, for each original input sequence we create two corrupted versions, each by independently picking some random positions to mask out and filling the masked positions with samples from the generator, just like how we do in token-level discrimination as described above. The two corruptions of a same original sequence $\mathbf{x}$, denoted as $\mathbf{x}^R_{i}$ and $\mathbf{x}^R_{j}$, are taken as a positive pair, and corruptions of other sequences within the same minibatch as $\mathbf{x}$ are regarded as negative examples, the set of which is denoted as $N(\mathbf{x})$. The CSP task is then to identify $\mathbf{x}^R_{j}$ in $N(\mathbf{x})$ for a given $\mathbf{x}^R_{i}$, and the contrastive loss is accordingly defined as:
\begin{equation}\label{eq:scl-loss}
\small
\mathcal{L}_{\rm CSP} \mspace{-1mu} (\mspace{-1mu} \mathbf{x}, \! \mathbf{x}^R_{i} \! , \! \mathbf{x}^R_{j} \mspace{-1mu} ; \!D \mspace{-1mu}) \!=\! - \log \! \frac{\exp \!\mspace{-1mu} \big(\mspace{-1mu} {\rm s}(\mathbf{x}^R_i \mspace{-1mu} , \mspace{-1mu} \mathbf{x}^R_j) / \tau \mspace{-1mu}\big)}{\sum_{\mathbf{x}\mspace{-1mu}^R_k \mspace{-1mu} \in \mspace{-1mu} N\!(\mspace{-1mu}\mathbf{x}\mspace{-1mu})} \mspace{-1mu} \exp \!\mspace{-1mu} \big(\mspace{-1mu} {\rm s}(\mathbf{x}^R_i \mspace{-1mu} , \mspace{-1mu} \mathbf{x}^R_k) / \tau \mspace{-1mu}\big)},\!\!\!
\end{equation}
where ${\rm s}(\cdot, \cdot)$ is the similarity measure between two sequences and $\tau$ is a temperature hyperparameter. We represent each sequence by the $\ell_2$-normalized representation of its {\tt [CLS]} token, {\it i.e.}, $\mu_D(\cdot)=$ $h_D(\cdot)_1 / \| h_D(\cdot)_1 \|$ where $h_D(\cdot)_1$ stands for the representation of the first token in a sequence output by the discriminator $D$, and determine the similarity as ${\rm s}(\mathbf{u}, \mathbf{v}) = \mu_D(\mathbf{u})^{\rm T} \mu_D(\mathbf{v})$. This contrastive learning task requires $\mathbf{x}^R_{i}$ and $\mathbf{x}^R_{j}$ to stay close to each other while away from other corrupted sequences in the same minibatch, and therefore encourages the discriminator to learn representations invariant to token-level alterations. A similar task has been considered recently by \citet{meng2021:coco} to help build general-domain PLMs, but it uses a different data transformation procedure to generate positive pairs by random cropping, resulting in asymmetric encoding of sequence pairs.

\subsection{Model Training}
Putting the generator and discriminator as well as their associated tasks together, we train eHealth by minimizing the following combined loss:
\begin{equation}\label{eq:loss}
\min_{G,D} \; \mathcal{L}_{\rm MLM} + \lambda_1 \mathcal{L}_{\rm RTD} + \lambda_2 \mathcal{L}_{\rm MTS} + \lambda_3 \mathcal{L}_{\rm CSP}.
\end{equation}
The first term is a generator loss, and the latter three are discriminator losses which are not propagated through the generator. $\lambda_1, \lambda_2, \lambda_3$ are hyperparameters balancing these loss terms. After pre-training, we throw out the generator and fine-tune only the discriminator on downstream tasks.
%We use the Adam stochastic optimizer \cite{kingma2014:adam} with linear learning rate warmup and decay. 

\section{Experiments}
This section first describes our experimental setups for pre-training and fine-tuning, and then presents evaluation results and further ablation.

\subsection{Pre-training Setups}\label{subsec:pretrain}
\paragraph{Pre-training Data.}
We use four Chinese datasets for pre-training: (i) {\it Dialogues} consisting of about 100 million de-identified doctor-patient dialogues from online healthcare services; (ii) {\it Articles} consisting of about 6.5 million popular scientific articles on medicine and healthcare oriented to the general public; (iii) {\it EMRs} consisting of about 6.5 million de-identified electronic medical records from specific hospitals; and ($\mspace{-1mu}$iv$\mspace{-1mu}$) {\it Textbooks} consisting of about 1,500 electronic textbooks on medicine and clinical pathology. The contents of these datasets are quite diversified, covering most aspects of bio-medicine, namely scientific, clinical, and consumer health \cite{jin2021:biomedical}. After collecting raw text, we conduct minimum pre-processing of deduplication and denoising on each of the four datasets. We then tokenize the text using a newly built in-domain vocabulary (detailed later). Sequences longer than 512 tokens are segmented into shorter chunks according to sentence boundaries, and those shorter than 32 tokens are discarded. Table~\ref{tab:pretrain-data} summarizes the datasets used for pre-training.

\paragraph{In-domain Vocabulary.}
Unlike previous studies that continually pre-train from and thereby use the vocabulary of a general-domain Chinese BERT, we train eHealth {\it from scratch} with its own in-domain vocabulary built specifically for Chinese biomedical text. \citet{gu2020:pubmedbert} have shown that training from scratch with an in-domain vocabulary is a better choice than continue pre-training while building English biomedical PLMs, primarily because the in-domain vocabulary can better handle highly specialized biomedical terms. This, however, has never been investigated in the Chinese biomedical field. To build the in-domain vocabulary, we randomly sample 1M documents from the pre-training data, convert all characters to lowercase, normalize special Unicodes like half-width characters or enclosed alphanumerics, and split Chinese characters, digits, and emoji Unicodes. Then we use the open-source implementation from the Tensor2Tensor library\footnote{\url{https://github.com/tensorflow/tensor2tensor}} to create a WordPiece vocabulary \cite{wu2016:wordpiece}. We throw out tokens appearing less than 5 times and keep the vocabulary of size to about 20K tokens, which is similar to the general-domain Chinese BERT. Table~\ref{tab:vocab} compares tokenization results obtained by (i) the original vocabulary of standard BERT and (ii) our newly built in-domain vocabulary. We can see that as both the two vocabularies are mainly based on single Chinese characters, the differences between them are not that significant as in English. But still the in-domain vocabulary works pretty better on abbreviations of specialized biomedical terms, including not only those rare ones like IHC (immunohistochemistry) and TSHR (thyroid stimulating hormone receptor), but also those relatively popular ones like AIDS (acquired immune deficiency syndrome) and MRI (magnetic resonance imaging). 

\begin{table}[t]
\small\centering%\setlength{\tabcolsep}{1pt}
\begin{tabular*}{0.48 \textwidth}{@{\extracolsep{\fill}}lccc}
\toprule
{\bf Corpus} & {\bf Size} & {\bf \# Tokens} & {\bf Sub-domain} \\
\midrule
Dialogues & ~~94.6GB   & 31.1B    & consumer health \\
Articles     & ~~11.2GB    & ~~3.5B & consumer health \\
EMRs       & ~~16.0GB    & ~~4.5B & clinical \\
Textbooks & ~~~~5.1GB & ~~1.6B & scientific \\
\midrule
Total         & 126.9GB        & 40.7B     & N/A \\
\bottomrule
\end{tabular*}
\caption{\label{tab:pretrain-data} Corpora used for eHealth pre-training.}
\end{table}

\begin{table}[t]
\small\centering\setlength{\tabcolsep}{2pt}
\begin{tabular*}{0.48 \textwidth}{@{\extracolsep{\fill}}rrl}
\toprule
\multicolumn{3}{@{}l}{免疫组化IHC测定TSHR阳性 (Positive expression of TSHR} \\
\multicolumn{3}{@{}l}{by immunohistochemistry (IHC))} \\
& BERT:    & 免, 疫, 组, 化, {\bf i}, {\bf \#\#hc}, 测, 定, {\bf ts}, {\bf \#\#hr}, 阳, 性 \\
& eHealth: & 免, 疫, 组, 化, {\bf ihc}, 测, 定, {\bf tshr}, 阳, 性 \\
\midrule
\multicolumn{3}{@{}l}{ECOG评分4分者 (Those with ECOG score of 4)} \\
& BERT: & {\bf eco}, {\bf \#\#g}, 评, 分, 4, 分, 者 \\
& eHealth: & {\bf ecog}, 评, 分, 4, 分, 者 \\
\midrule
\multicolumn{3}{@{}l}{但不包括HIV/AIDS (But excluding HIV/AIDS)} \\
& BERT: & 但, 不, 包, 括, hiv, /, {\bf ai}, {\bf \#\#ds} \\
& eHealth: & 但, 不, 包, 括, hiv, /, {\bf aids} \\
\midrule
\multicolumn{3}{@{}l}{胸部增强CT及头颅MRI (Enhanced chest CT \& skull MRI)} \\
& BERT: & 胸, 部, 增, 强, ct, 及, 头, 颅, {\bf mr}, {\bf \#\#i} \\
& eHealth: & 胸, 部, 增, 强, ct, 及, 头, 颅, {\bf mri} \\
\bottomrule
\end{tabular*}
\caption{\label{tab:vocab} Comparison of tokenization results obtained by BERT and eHealth. Differences highlighted in bold.}
\end{table}

\begin{table*}[t]
\small\centering%\setlength{\tabcolsep}{1pt}
\begin{tabular*}{1 \textwidth}{@{\extracolsep{\fill}}llrrrl}
\toprule
{\bf Dataset} & {\bf Task} & {\bf Train} & {\bf Dev} & {\bf Test} & {\bf Metric} \\
\midrule
CMeEE         & Named Entity Recognition    & 15,000 & 5,000 & 3,000   & Micro-F1 \\
CMeIE           & Relation Extraction               & 14,339 & 3,585 & 4,482   & Micro-F1 \\
CHIP-CDN    & Clinical Term Normalization  & 6,000   & 2,000 & 10,192 & Micro-F1 \\
CHIP-CTC     & Sentence Classification        & 22,962 & 7,682 & 10,000 & Macro-F1 \\
KUAKE-QIC  & Sentence Classification        & 6,931   & 1,955 & 1,994   & Accuracy \\
CHIP-STS     & Sentence Pair Matching       & 16,000 & 4,000 & 10,000 & Macro-F1 \\
KUAKE-QTR & Sentence Pair Matching       & 24,174 & 2,913 & 5,465   & Accuracy \\
KUAKE-QQR & Sentence Pair Matching      & 15,000 & 1,600 & 1,596   & Accuracy \\
\midrule
cMedQNLI \cite{zhang2020:mcbert}    & Question Answer Matching & 80,950   & 9,065   & 9,969   & Micro-F1 \\
webMedQA \cite{he2019:webmedqa}  & Question Answer Matching & 252,850 & 31,605 & 31,655 & Precision@1 \\
NLPEC \cite{li2020:nlpec}                    & Multiple Choice                   & 18,117   & 2,500   & 550       & Accuracy \\
\bottomrule
\end{tabular*}
\caption{\label{tab:datasets} Downstream tasks used for evaluation. Tasks in the first group are from CBLUE \cite{zhang2021:cblue}.}
\end{table*}

\paragraph{Pre-training Configurations.}
We train eHealth with the standard {\it base-size} configuration, just like most previous biomedical PLMs. The discriminator gets 12 Transformer layers, each with 12 attention heads, 768 hidden size, and 3072 intermediate size. And we follow \citet{clark2020:electra} to set the generator 1/3 the size of the discriminator and tie their token and positional embeddings. To generate masked positions, we perform Chinese word segmentation and use the whole word masking strategy \cite{cui2020:macbert}. We also use dynamic masking with masked positions decided on-the-fly. During pre-training, we mostly follow the hyperparameters recommended by ELECTRA and do not conduct hyperparameter tuning. For newly introduced hyperparameters, we set the loss balancing terms $\lambda_1$ $=50$, $\lambda_2=20$, $\lambda_3$ $=1$ (cf. Eq.~(\ref{eq:loss})), the number of sampled non-original tokens $k=5$ (cf. Eq.~(\ref{eq:mts-prob})), and temperature $\tau=0.07$ (cf. Eq.~(\ref{eq:scl-loss})). We train with a batch size of 384 and max sequence length of 512 for 1.65M steps. The full set of pre-training hyperparameters is listed in Appendix~\ref{appendix:pretraining}.

\subsection{Evaluation Setups}
\paragraph{Downstream Tasks.} We evaluate on the Chinese Biomedical Language Understanding Evaluation ({\it CBLUE}) benchmark \cite{zhang2021:cblue}, which is composed of 8 diversified biomedical NLP tasks, ranging from medical text classification and matching to medical information extraction and medical term normalization. We further consider three medical question answering tasks, namely {\it cMedQNLI} \cite{zhang2020:mcbert}, {\it webMedQA} \cite{he2019:webmedqa}, and {\it NLPEC} \cite{li2020:nlpec}. The former two are formalized as question-answer matching problems, and the last one a multiple choice problem. Table~\ref{tab:datasets} summarizes the train, dev, test split and metric used for each task. We refer readers to Appendix~\ref{appendix:cblue} and \ref{appendix:medical-qa} for further details.

\paragraph{Baseline Models.}
We compare eHealth against state-of-the-art general-domain Chinese PLMs of: (i) {\it BERT-base} \cite{devlin2019:bert}; (ii) {\it ELECTRA-base/large} \cite{clark2020:electra}; (iii) {\it RoBERTa-wwm} {\it -ext-base/large} \cite{liu2019:roberta} trained via MLM with whole word masking strategy; (iv) {\it MacBERT-base/large} \cite{cui2020:macbert} trained via improved MLM as a correction task. BERT-base is officially released by Google,\footnote{\url{https://github.com/google-research/bert}} and the other models are released by \citet{cui2020:macbert}.\footnote{\url{https://github.com/ymcui/MacBERT}} Besides, we compare to Chinese biomedical PLMs including: (v) {\it PCL-MedBERT};\footnote{\url{https://code.ihub.org.cn/projects/1775}} (vi) {\it MC-BERT} \cite{zhang2020:mcbert};\footnote{\url{https://github.com/alibaba-research/ChineseBLUE}} (vii) {\it EMBERT} \cite{cai2021:embert}; and (viii) {\it SMedBERT} \cite{zhang2021:smedbert}, all initialized from Google's BERT-base. The full models of EMBERT and SMedBERT are not released to the public, so we just copy the results reported by their authors on medical question answering tasks.

\paragraph{Fine-tuning Configurations.} $\!\!$During fine-tuning, we build a lightweight task-specific head on top of the pre-trained encoders for each task. The specific design of these heads is elaborated in Appendix~\ref{appendix:heads}. For each PLM on each task, we tune the batch size, learning rate, and training epochs in their respective ranges, and determine the optimal setting according to dev performance averaged over three runs with different seeds. The other hyperparameters are set to their default values as in ELECTRA \cite{clark2020:electra}. The full set of fine-tuning hyperparameters is listed in Appendix~\ref{appendix:finetuning}.

\begin{table*}[t]
\small\centering%\setlength{\tabcolsep}{1pt}
\begin{tabular*}{1 \textwidth}{@{\extracolsep{\fill}}lccccccccc}
\toprule
{\bf Model} & {\bf CMeEE} & {\bf CMeIE} & {\bf CDN} & {\bf CTC} & {\bf STS} & {\bf QIC} & {\bf QTR} & {\bf QQR} & {\bf Avg.} \\
%\midrule
%Human Performance        & 67.0 & 66.0 & 65.0 & 78.0 & 93.0 & 88.0 & 71.0 & 89.0 & 77.1 \\
\midrule
\multicolumn{10}{@{}l}{\it General-domain base-sized models} \\
BERT-base                       & 66.5 & 60.6 & 69.7 & 68.6 & 84.7 & 85.2 & 59.2 & 82.5 & 72.1 \\
ELECTRA-base                & 65.1 & 60.4 & 69.9 & 67.7 & 84.4 & 85.2 & 61.8 & 84.0 & 72.3 \\
MacBERT-base                & 66.8 & 61.5 & 69.7 & 69.1 & 84.4 & 86.0 & 61.0 & 83.5 & 72.7 \\
RoBERTa-wwm-ext-base & 66.7 & 61.4 & 69.3 & 68.3 & 84.2 & 86.0 & 60.9 & 82.7 & 72.4 \\
%ERNIE2.1-base                & 67.1 & 61.2 & 70.2 & 69.2 & 85.5 & 84.9 & 62.0 & 85.2 & 73.2 \\
%ERNIE2.2-base                & 66.8 & 61.0 & 70.4 & 70.8 & 85.7 & 84.3 & 62.9 & 85.0 & 73.4 \\
%ERNIE2.3-base                & 67.1 & 60.7 & 70.2 & 68.5 & 86.1 & 85.1 & 62.2 & 85.0 & 73.1 \\
\midrule
\multicolumn{10}{@{}l}{\it General-domain large-sized models} \\
ELECTRA-large               & 66.1 & 59.3 & 70.8 & 68.9 & 85.1 & 84.1 & 62.0 & 85.7 & 72.8 \\
MacBERT-large                & \underline{67.6} & \underline{62.2} & \underline{70.9} & 69.7 & \underline{86.5} & 85.7 & \underline{62.5} & 83.5 & 73.6 \\
RoBERTa-wwm-ext-large & 67.3 & \underline{62.2} & 70.6 & \underline{70.6} & 85.4 & \underline{86.7} & 61.7 & \underline{86.1} & \underline{73.8} \\
%ERNIE2.1-large                & 67.3 & 61.4 & 70.0 & 69.4 & 86.0 & 85.2 & 63.7 & 84.8 & 73.5 \\
\midrule
\multicolumn{10}{@{}l}{\it Biomedical base-sized models} \\
MC-BERT-base               & 66.6 & 60.7 & 70.1 & 69.1 & 85.4 & 85.3 & 61.6 & 82.3 & 72.6 \\
PCL-MedBERT-base       & 66.6 & 60.8 & 69.9 & {\bf 70.4} & 84.8 & 85.3 & 60.2 & 83.3 & 72.7 \\
%eHealth-base (ours)        & {\bf 67.0} & {\bf 61.6} & {\bf 71.2} & {\bf 70.5} & {\bf 86.2} & {\bf 87.4} & {\bf 63.5} & {\bf 85.5} & {\bf 74.1} \\
eHealth-base (ours)        & {\bf 66.9} & {\bf 62.1} & {\bf 71.9} & 69.3 & {\bf 86.2} & {\bf 87.3} & {\bf 63.9} & {\bf 85.7} & {\bf 74.2} \\
%ERNIE-Health-base     & 66.5 & 60.9 & 71.3 & 71.5 & 86.6 & 87.1 & 63.3 & 85.6 & 74.1 \\
\bottomrule
\end{tabular*}
\caption{\label{tab:cblue-test} Performance (\%) of different PLMs on CBLUE test sets. Results generated by the single best run on dev sets. Best scores from {\bf base-sized} models highlighted in bold, and best scores from \underline{large-sized} models underlined.}
\end{table*}

\begin{table}[t]
\small\centering\setlength{\tabcolsep}{1pt}
\begin{tabular*}{0.48 \textwidth}{@{\extracolsep{\fill}}lccc}
\toprule
& {\bf cMedQNLI} & {\bf webMedQA} &{\bf NLPEC} \\
{\bf Model} & dev | test & dev | test & dev | test \\
\midrule
\multicolumn{4}{@{}l}{\it General-domain base-sized models} \\
BERT-base        & 96.4 | 96.4 & 79.6 | 79.8 & 67.1 | 54.6 \\
ELECTRA-base & 96.0 | 95.9 & 79.2 | 79.1 & 69.8 | 54.1 \\
MacBERT-base  & 96.3 | 96.2 & 79.9 | 79.8 & 68.7 | 53.8 \\
RoBERTa-base  & 96.2 | 96.2 & 79.7 | 79.9 & 68.1 | 54.3 \\
\midrule
\multicolumn{4}{@{}l}{\it General-domain large-sized models} \\
ELECTRA-large & \underline{96.4} | 96.2 & \underline{80.0} | 80.1                    & \underline{71.8} | \underline{60.0} \\
MacBERT-large  & 96.3 | \underline{96.3} & \underline{80.0} | \underline{80.4} & 70.8 | 56.7 \\
RoBERTa-large  & 96.3 | 96.2                   & 79.7 | 79.7                                      & 71.1 | 56.5 \\
\midrule
\multicolumn{4}{@{}l}{\it Biomedical base-sized models} \\
MC-BERT-base           & 96.4 | 96.5              & 80.0 | 79.9         & 68.2 | 54.2 \\
PCL-MedBERT-base   & 96.3 | 96.2              & 79.2 | 79.5         & 67.4 | 52.0 \\
EMBERT$^\dag$       & ~~~--~~ | 96.6          & ~~~--~~ | 80.6   & ~~~--~~ | ~~~--~~ \\
SMedBERT$^\ddag$ & 96.6 | 96.9                & 79.3 | {\bf 81.7} & ~~~--~~ | ~~~--~~ \\
eHealth-base (ours)   & {\bf 97.3} | {\bf 97.2} & {\bf 80.5} | 80.7 & {\bf 73.6} | {\bf 62.4} \\
\bottomrule
\end{tabular*}
\caption{\label{tab:qa-tasks} Performance (\%) of different PLMs on medical QA tasks. RoBERTa-base/large refers to RoBERTa-wwm-ext-base/large. Results marked by $\dag$ and $\ddag$ copied from original literatures \cite{cai2021:embert,zhang2021:smedbert}. Other results produced by ourselves, averaged over best three runs on the dev set of each task. Best scores from {\bf base-sized} models highlighted in bold and best scores from \underline{large-sized} models underlined.}
\end{table}

\subsection{Main Results}
Table~\ref{tab:cblue-test} reports the performance of different PLMs on CBLUE test sets. Note that CBLUE test labels are not released, and one has to submit prediction files to retrieve final scores. To avoid frequent submissions that probe the unseen test labels, we only submit best single run on dev sets for testing. The results show that: (i) The two previous biomedical PLMs, MC-BERT and PCL-MedBERT, indeed perform better than general-domain BERT-base from which they started continual pre-training, verifying the effectiveness of domain adaptation in building domain-specific language models. However, these two biomedical PLMs fail to surpass some more advanced general-domain PLMs, {\it e.g.}, MacBERT, of the same model size. (ii) As the model size increases, general-domain large-sized PLMs perform better than those base-sized, {\it e.g.}, ELECTRA-large, MacBERT-large, and RoBERTa-wwm-ext-large obtain averaged improvements of 0.5\%, 0.9\%, and 1.4\% respectively over their base-sized models. (iii) eHealth, as a base-sized biomedical PLM, outperforms all baseline PLMs in terms of average score, no matter those from the general or biomedical domain, and no matter those base-sized or large-sized. It achieves an average improvement of 1.5\% over PCL-MedBERT-base, {\it i.e.}, the best performing direct opponent of the same model size, and even that of 0.4\% over the best performing large-sized model RoBERTa-wwm-ext-large. These results demonstrate the effectiveness and superiority of eHealth in biomedical text understanding.

Table~\ref{tab:qa-tasks} further reports the performance of these PLMs on medical question answering tasks, where scores are averaged over the best three runs selected on the dev split for each task. From the results we can observe similar phenomena as on the CBLUE benchmark. Still eHealth consistently outperforms almost all those PLMs, showing its superior ability in medical question answering.

\begin{table*}[t]
\small\centering%\setlength{\tabcolsep}{2pt}
\begin{tabular*}{1 \textwidth}{@{\extracolsep{\fill}}lccccccccc}
\toprule
{\bf Model} & {\bf CMeEE} & {\bf CMeIE} & {\bf CDN} & {\bf CTC} & {\bf STS} & {\bf QIC} & {\bf QTR} & {\bf QQR} & {\bf Avg.} \\
\midrule
The full setting   & {\bf 66.56} & {\bf 61.62} & \underline{70.29} & \underline{69.58} & \underline{85.13} & {\bf 87.46} & {\bf 62.00} & {\bf 85.53} & {\bf 73.52} \\
\hspace{6pt}w/o CSP  & \underline{66.47} & \underline{61.25} & 69.81 & {\bf 69.65} & 84.61 & \underline{86.71} & \underline{61.54} & \underline{84.52} & \underline{73.07} \\
\hspace{6pt}w/o MTS  & 65.76 & 60.23 & {\bf 70.43} & 68.06 & {\bf 85.44} & 85.61 & 61.36 & 84.34 & 72.65 \\
\hspace{6pt}w/o CSP \& MTS & 65.56 & 60.01 & 70.08 & 68.46 & 84.35 & 86.51 & 61.08 & 84.40 & 72.56 \\
\midrule\midrule
R weights + B vocab & {\bf 66.56} & {\bf 61.62} & \underline{70.29} & {\bf 69.58} & \underline{85.13} & {\bf 87.46} & 62.00 & \underline{85.53} & {\bf 73.52} \\
E weights + E vocab & 65.92 & \underline{61.54} & {\bf 70.86} & \underline{69.53} & {\bf 85.75} & 86.21 & {\bf 62.38} & {\bf 85.59} & \underline{73.47} \\
R weights + E vocab & \underline{66.33} & 61.06 & 70.19 & 69.50 & 84.32 & \underline{87.31} & \underline{62.33} & 85.40 & 73.30 \\
\bottomrule
\end{tabular*}
\caption{\label{tab:ablation} Effects of pre-training tasks (top) and initialization strategies (bottom) on CBLUE test sets, where results are generated by single best run on dev sets. All variants are base-sized, trained with batch size 128 for 500K steps. R/B/E in the bottom group stands for R(andom)/B(iomedical)/E(LECTRA), respectively. Within each group {\bf best} scores are highlighted in bold, and \underline{second best} scores underlined.}
\end{table*}

\subsection{Ablation Studies}
We provide ablation studies on CBLUE benchmark to show the effects of different pre-training tasks and initialization strategies in eHealth. All variants below are base-sized, trained with the same setting as described in Section~\ref{subsec:pretrain}. The only exception is that we train with a smaller batch size of 128 for only 500K steps. 

\paragraph{Effects of Pre-training Tasks.}
$\!\!$The discriminator of eHealth is trained in a multi-task fashion, {\it i.e.}, (i) token-level discrimination of RTD and MTS and (ii) sequence-level discrimination of CSP. To investigate the effects of different pre-training tasks, we make comparison among: (i) {\it the full setting} where the discriminator is trained via RTD, MTS, and CSP; (ii) {\it w/o CSP} where the sequence-level CSP is removed; (iii) {\it w/o MTS} where the token-level MTS is removed; and (iv) {\it w/o CSP \& MTS} where both CSP and MTS are removed and thus degenerates to standard ELECTRA pre-training. Table~\ref{tab:ablation} (top) lists the results on CBLUE benchmark, from which we can see that: (i) The full setting performs the best among the four variants, always reporting the best or second best scores on all the 8 diversified tasks. Compared to standard ELECTRA pre-training (w/o CSP \& MTS), it achieves an average improvement of 0.96\%. This demonstrates the usefulness of our pre-training tasks, in particular CSP and MTS, to build effective PLMs. (ii) No matter CSP or MTS, when applied alone, is able to improve the standard ELECTRA pre-training solely with RTD. Between the two tasks, MTS is, in general, more powerful than CSP. Removing MTS brings an average drop of 0.87\% on CBLUE test sets, while removing CSP only brings that of 0.45\% on the same benchmark.

\paragraph{Effects of Initialization Strategies.}
In this work we train eHealth entirely from scratch, with an in-domain vocabulary built specifically for Chinese biomedical text and the model weights randomly initialized. We refer to this strategy as ``{\it R(andom) weights + B(iomedical) vocab}''. We compare it to the widely adopted continue pre-training strategy, where model weights are initialized from a general-domain ELECTRA and the associated vocabulary is also used, referred to as ``{\it E(LECTRA) weights + E(LECTRA) vocab}''. Besides, to further verify the effects of that in-domain vocabulary, we consider another setting ``{\it R(andom) weights + E(LECTRA) vocab}'', where model weights are still randomly initialized but the ELECTRA vocabulary is used. Table~\ref{tab:ablation} (bottom) lists the results on CBLUE benchmark, from which we can see that: (i) Pre-training from scratch with the newly built in-domain vocabulary (R weights + B vocab) overall performs better than continue pre-training (E weights + E vocab), even under a relatively small number of training steps up to 500K.\footnote{The advantage, in fact, will be expanded further as the training step increases according to our initial experiments.} (ii) The improvements mainly come from the in-domain vocabulary. After replacing the vocabulary with that of the general-domain ELECTRA (R weights + E vocab), the overall performance drops from 73.52\% to 73.30\%.

%In this work we train eHealth entirely from scratch, with an in-domain vocabulary built specifically for Chinese biomedical text and the model weights randomly initialized. We refer to this strategy as {\it fully from scratch} and compare it to the widely used {\it continue pre-training}, where model weights are initialized from the general-domain ELECTRA and the associated vocabulary is also used. Besides, to further verify the effects of that in-domain vocabulary, we consider another setting of {\it partially from scratch}, where model weights are still randomly initialized but the ELECTRA vocabulary is used. Table~\ref{tab:ablation} (bottom) presents the results on CBLUE benchmark, from which we can see that: (i) 

\section{Conclusion}
This work presents eHealth, a Chinese biomedical language model pre-trained from in-domain text of de-identified online doctor-patient dialogues, electronic medical records, and textbooks. Unlike most previous studies that directly adapt general-domain PLMs to the biomedical domain, eHealth is trained from scratch with a new self-supervised generator-discriminator framework. The generator is used to produce corrupted input and is discarded after pre-training. The discriminator, as the final encoder, is trained via multi-level discrimination: (i) token-level discrimination that detects input tokens corrupted by the generator and selects original tokens from plausible candidates; and (ii) sequence-level discrimination that further detects corruptions of a same original sequence from those of the others. As such, eHealth can learn language semantics at both levels. Experimental results on CBLUE and 3 medical QA benchmarks demonstrate the effectiveness and superiority of eHealth, which consistently outperforms state-of-the-art PLMs from both the general and biomedical domains. We release our pre-trained model to the public, which could be applied rather easily during fine-tuning. 
%We release our pre-trained model to the public. As the model relies solely on text, it could be applied rather easily during fine-tuning. 

\newpage

\bibliography{acl2022}
\bibliographystyle{acl_natbib}

\appendix

\section{Pre-training Hyperparameters}\label{appendix:pretraining}
We mostly use the same hyperparameters as ELECTRA \cite{clark2020:electra} and do not conduct hyperparameter tuning during pre-training. As for those newly introduced hyperparameters, we sample $k=$ $5$ non-original tokens for a certain position in the MTS task, use a temperature $\tau=0.07$ in the CSP task, and set the loss balancing tradeoffs $\lambda_1=50$, $\lambda_2=20$, $\lambda_3=1$. The full pre-training setting is listed in Table~\ref{tab:pretrain-params}.

\section{Fine-tuning Hyperparameters}\label{appendix:finetuning}
During fine-tuning, we mostly use the default setting as suggested by BERT \cite{devlin2019:bert} and ELECTRA \cite{clark2020:electra}, listed in Table~\ref{tab:finetune-params}. We also use exponential moving average (EMA) with a decay coefficient $\alpha$ of $0.9999$. Then for each task we specify a proper maximum sequence length, tune for each PLM the batch size, learning rate, and training epochs in their respective ranges, and determine optimal configurations according to dev performance. The full tuning ranges are listed in Table~\ref{tab:ranges}.

\begin{table}[t]
\small\centering%\setlength{\tabcolsep}{1pt}
\begin{tabular*}{0.48 \textwidth}{@{\extracolsep{\fill}}ll}
\toprule
{\bf Hyperparameter} & {\bf Value} \\
\midrule
Number of Layers               & 12 \\
Hidden size                         & 768 \\
Intermediate size                & 3072 \\
Number of attention heads & 12 \\
Attention head size             & 64 \\
Embedding size                  & 768 \\
Generator size (multiplier for hidden size,  & \multirow{2}{*}{1/3} \\
intermediate size, number of attention heads) & \\
Mask percentage     & 15 \\
Learning rate decay & Linear \\
Warmup steps          & 10000 \\
Learning rate            & 2e-4 \\
Adam $\epsilon$      & 1e-6 \\
Adam $\beta_1$      & 0.9 \\
Adam $\beta_2$      & 0.999 \\
Attention dropout     & 0.1 \\
Dropout                    & 0.1 \\
Weight decay           & 0.01 \\
Max sequence length & 512 \\
Batch size                  & 384 \\
Training steps            & 1.65M \\
Loss tradeoff $\lambda_1$ & 50 \\
Loss tradeoff $\lambda_2$ & 20 \\
Loss tradeoff $\lambda_3$ & 1 \\
Multi-token selection $k$  & 5 \\
Contrastive sequence prediction $\tau$ & 0.07 \\
\bottomrule
\end{tabular*}
\caption{\label{tab:pretrain-params} Pre-training hyperparameters.}
\end{table}

\begin{table}[t]
\small\centering%\setlength{\tabcolsep}{1pt}
\begin{tabular*}{0.48 \textwidth}{@{\extracolsep{\fill}}ll}
\toprule
{\bf Hyperparameter} & {\bf Value} \\
\midrule
Learning rate decay & Linear \\
Warmup ratio           & 0.1 \\
Adam $\epsilon$      & 1e-8 \\
Adam $\beta_1$      & 0.9 \\
Adam $\beta_2$      & 0.999 \\
Attention dropout     & 0.1 \\
Dropout                    & 0.1 \\
Weight decay           & 0.01 \\
EMA decay              & 0.9999 \\
\bottomrule
\end{tabular*}
\caption{\label{tab:finetune-params} Default fine-tuning hyperparameters.}
\end{table}

\begin{table*}[t]
\small\centering%\setlength{\tabcolsep}{1pt}
\begin{tabular*}{1 \textwidth}{@{\extracolsep{\fill}}llllr}
\toprule
{\bf Task} & {\bf Batch size} & {\bf Learning rate} & {\bf Epochs} & {\bf Length} \\
\midrule
\multicolumn{5}{@{}l}{\it CBLUE benchmark} \\
CMeEE          & 32 & 6e-5, 1e-4 & 2, 4, 8, 12 & 128 \\
CMeIE            & 12 & 6e-5 & 50, 100, 150, 200, 250 & 300 \\
CHIP-CDN     & 256 & 3e-5, 6e-5, 1e-4 & 2, 4, 8, 12, 16 & 32 \\
CHIP-CTC      & 8, 16, 32 & 3e-5, 6e-5, 1e-4 & 2, 4, 8, 12, 16 & 160 \\
CHIP-STS      & 8, 16, 32 & 3e-5, 6e-5, 1e-4 & 2, 4, 8, 12, 16 & 96 \\
KUAKE-QIC   & 8, 16, 32 & 3e-5, 6e-5, 1e-4 & 2, 4, 8, 12, 16 & 128 \\
KUAKE-QTR  & 8, 16, 32 & 3e-5, 6e-5, 1e-4 & 2, 4, 8, 12, 16 & 64 \\
KUAKE-QQR & 8, 16, 32 & 3e-5, 6e-5, 1e-4 & 2, 4, 8, 12, 16 & 64 \\
\midrule
\multicolumn{5}{@{}l}{\it Medical QA tasks} \\
cMedQNLI   & 8, 16, 32 & 3e-5, 6e-5, 1e-4 & 1, 2, 3, 4 & 512 \\
webMedQA & 16, 32, 64 & 1e-5, 2e-5, 3e-5 & 1, 2, 3, 4 & 512 \\
NLPEC        & 32           & 2e-5, 3e-5, 6e-5 & 10, 20, 30, 40 & 512 \\
\bottomrule
\end{tabular*}
\caption{\label{tab:ranges} Hyperparameter tuning ranges on CBLUE and medical QA benchmarks.}
\end{table*}

\section{CBLUE Benchmark}\label{appendix:cblue}
CBLUE \cite{zhang2021:cblue}\footnote{\url{https://github.com/CBLUEbenchmark/CBLUE}} is a benchmark for Chinese biomedical language understanding evaluation, consisting of 8 diversified biomedical NLP tasks as follows.

\paragraph{CMeEE:} $\!\!\!\!$Chinese Medical Entity Extraction.\footnote{\url{http://www.cips-chip.org.cn/2020/eval1}} The task is to identify medical entities from a given sentence and classify the entities into nine categories including disease, symptom, drug, etc. The dataset contains 15K/5K/3K train/dev/test examples from textbooks of clinical pediatrics. 

\paragraph{CMeIE:} $\!\!\!\!$Chinese Medical Information Extraction \cite{guan2020:cmeie}.\footnote{\url{http://www.cips-chip.org.cn/2020/eval2}} The task is to recognize both medical entities and their relationships from a given sentence according to a predefined schema. There are 44 relations defined in the schema, along with their subject/object entity types. The dataset contains about 14K/3.5K/4.5K train/dev/test examples, which are also from textbooks of clinical pediatrics.

\paragraph{CHIP-CDN:} $\!\!\!\!$CHIP Clinical Diagnosis Normalization.\footnote{\url{http://www.cips-chip.org.cn/2020/eval3}} The task is to normalize original diagnostic terms into standard terminologies from the International Classification of Diseases (ICD-10), Beijing Clinical Edition v601. The dataset contains about 6K/2K/10K train/dev/test examples collected from de-identified electronic medical records.

\paragraph{CHIP-CTC:} $\!\!\!\!$CHIP Clinical Trial Classification \cite{zong2021:chipctc}.\footnote{\url{https://github.com/zonghui0228/chip2019task3}} The task is to categorize eligibility criteria of clinical trials into 44 predefined semantic classes including age, disease, symptom, etc. The dataset consists of about 23K/7.5K/10K train/dev/test examples collected from the website of Chinese Clinical Trial Registry.%\footnote{\url{https://www.chictr.org.cn/}}

\paragraph{CHIP-STS:} $\!\!\!\!$CHIP Semantic Textual Similarity.\footnote{\url{http://www.cips-chip.org.cn:8000/evaluation}} The task is to identify whether the semantics of two medical questions are identical or not. The dataset contains 16K/4K/10K train/dev/test question pairs collected from online healthcare services, covering 5 diseases including diabetes, hypertension, hepatitis, aids, and breast cancer.

\paragraph{KUAKE-QIC:} $\!\!\!\!$KUAKE Query Intent Classification. The task is to classify the intent of a medical search query into one of 11 predefined categories like diagnosis, etiology analysis, medical advice, etc. The dataset contains about 7K/2K/2K queries in the train/dev/test split, collected from Alibaba QUAKE search engine.

\paragraph{KUAKE-QTR:} $\!\!\!\!$KUAKE Query Title Relevance. The task aims to estimate the relevance between a search query and a webpage title. The relevance is divided into four levels: perfectly match, partially match, slightly match, and mismatch. The dataset contains about 24K/3K/5.5K query-title pairs in the train/dev/test split, collected from Alibaba QUAKE search engine.

\paragraph{KUAKE-QQR:} $\!\!\!\!\!$KUAKE Query Query Relevance. Similar to KUAKE-QTR, the task is to estimate the relevance between two search queries $Q_1$ and $Q_2$. The relevance is divided into three levels: perfectly match, $Q_2$ is a subset of $Q_1$, $Q_2$ is a superset of $Q_1$ or mismatch. The dataset contains approximately 15K/1.6K/1.6K pairs of queries in the train/dev/test split. The queries are also collected from Alibaba QUAKE search engine.

\section{Medical QA Tasks}\label{appendix:medical-qa}
Besides CBLUE, we consider three medical question answering (QA) tasks, detailed as follows.

%namely cMedQNLI \cite{zhang2020:mcbert}, webMedQA \cite{he2019:webmedqa}, and NLPEC \cite{li2020:nlpec}

\paragraph{cMedQNLI:} $\!\!\!\!$This is a Chinese medical QA dataset which formalizes QA as a question answer matching problem \cite{zhang2020:mcbert}.\footnote{\url{https://github.com/alibaba-research/ChineseBLUE}} Given a question answer pair, the task is to identify whether the answer addresses the question or not. The dataset contains about 81K/9K/10K question answer pairs in the train/dev/test split.

\paragraph{webMedQA:} $\!\!\!\!$This dataset also formalizes medical QA as a question answer matching problem \cite{he2019:webmedqa},\footnote{\url{https://github.com/hejunqing/webMedQA}} just like cMedQNLI. But it is much larger, containing roughly 250K/31.5K/31.5K question answer pairs in the train/dev/test split.

\paragraph{NLPEC:} $\!\!\!\!$This is a multiple choice QA dataset constructed using simulated and real questions from the National Licensed Pharmacist Examination in China \cite{li2020:nlpec}.\footnote{\url{http://112.74.48.115:8157}} Given a question along with five answer candidates, the task is to select the most plausible answer from the candidates using textual evidences extracted from the official exam guide. The dataset contains about 18K/2.5K/0.5K questions in the train/dev/test split.

\section{Task-specific Heads}\label{appendix:heads}
We devise lightweight task-specific heads on top of pre-trained Transformer encoders to solve downstream tasks in various forms. These task-specific heads are roughly categorized into five groups, used for named entity recognition, relation extraction, single sentence classification, sentence pair classification, and multiple choice QA, respectively.

\begin{figure}[t]
\centering
\includegraphics[width=0.5\textwidth]{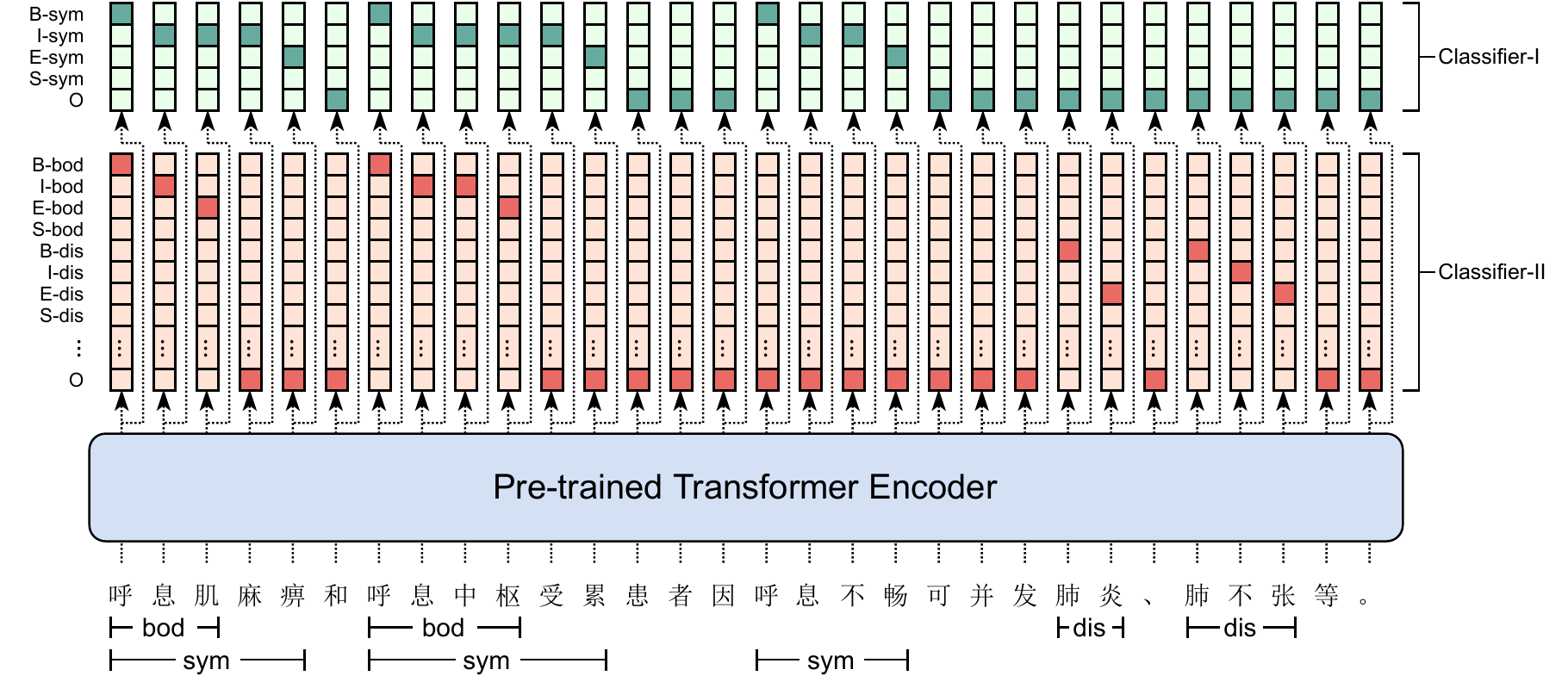}
\caption{A running example illustrating the sequence tagging head used for CMeEE. Dark shaded entries represent a ground truth label of 1, and light shaded entries a ground truth label of 0.}
\label{fig:cmeee}
\end{figure}

\paragraph{Named Entity Recognition.} CMeEE is the only task of this kind. It recognizes medical entities and classifies them into 9 predefined types. Nesting is allowed only in symptom entities, but not in entities of the other types. We therefore use a two-stream sequence tagging head for this task, one to identify symptom entities and the other to identify entities of the other 8 types. We choose the BIOES ({\it i.e.}, Be-gin, Inside, Outside, End, Single) tagging scheme \cite{ratinov2009:bioes}. Given a sequence with its contextualized representations output by a pre-trained encoder, we build two classifiers on top of these representations. The first assigns each token in the sequence into 5 classes to annotate symptom entities (4 type-specific {\tt B-}, {\tt I-}, {\tt E-}, {\tt S-} tags plus {\tt O} tag), while the second assigns it into 33 classes to annotate entities of other types (32 type-specific {\tt B-}, {\tt I-}, {\tt E-}, {\tt S-} tags plus {\tt O} tag). The two classifiers are trained jointly with a 1:1 balanced combined loss. Figure~\ref{fig:cmeee} gives a running example illustrating this two-stream sequence tagging head.

\begin{figure}[t]
\centering
\includegraphics[width=0.5\textwidth]{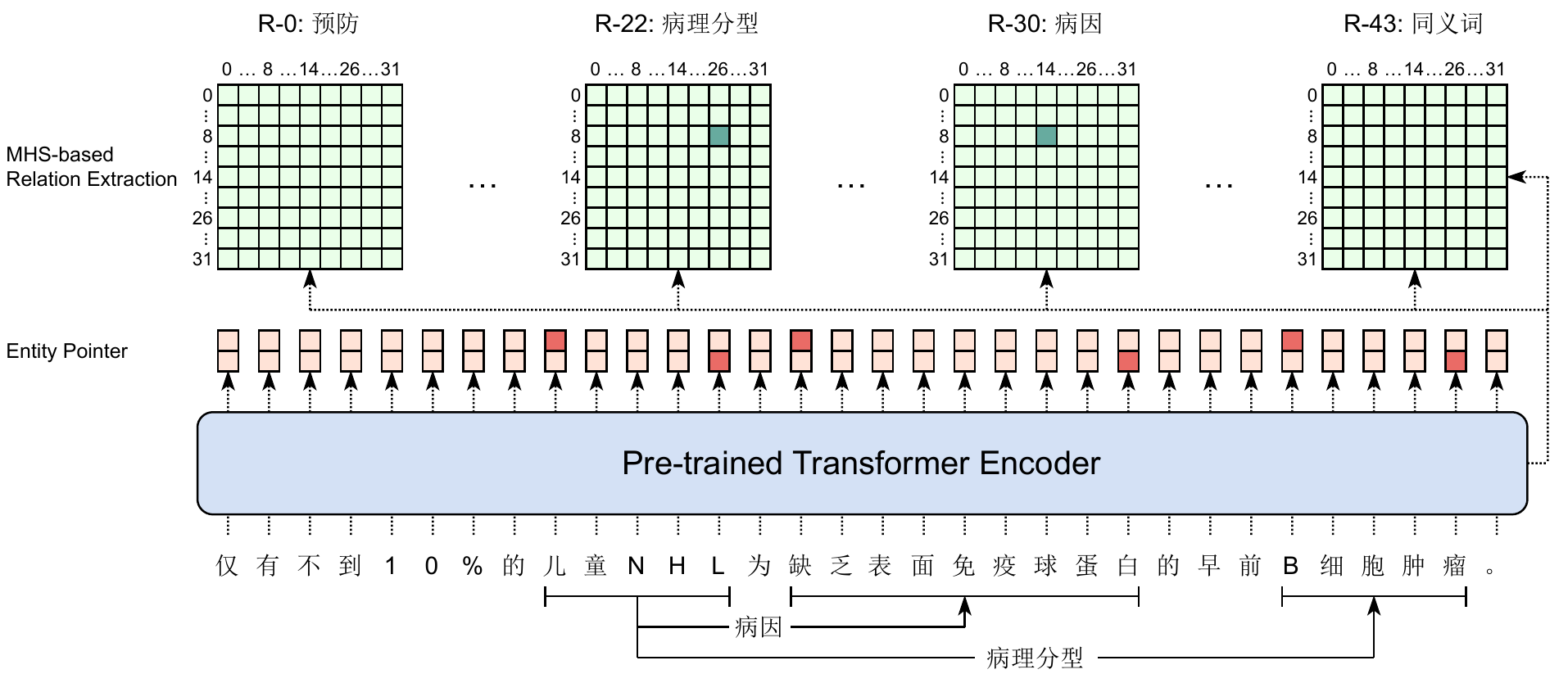}
\caption{A running example illustrating the multi-head selection layer used for joint entity and relation extraction in CMeIE. Dark shaded entries represent a ground truth label of 1, and light shaded entries a ground truth label of 0.}
\label{fig:cmeie}
\end{figure}

\paragraph{Relation Extraction.} CMeIE is the only task of this kind. It extracts subject-relation-object triples according to a predefined schema. There are totally 44 relations defined in the schema and overlapping is allowed between these relations, {\it i.e.}, one entity may belong to multiple triples of different relations \cite{zeng2018:overlap}. To solve this overlapping problem, we use a multi-head selection (MHS) layer for joint entity and relation extraction \cite{bekoulis2018:mhs}. As illustrated in Figure~\ref{fig:cmeie}, an entity pointer is adopted to identify start and end of entity spans, and then an MHS mechanism is further employed to recognize possible relationships between pairs of entity spans. The MHS module predicts if there exists a relation $k$ between a subject entity starting at position $i$ and an object entity starting at position $j$ for every $i$, $j$, and $k$. This prediction probability is calculated via a relation-specific biaffine operation imposed upon the starting token representations of subject and object entities. Finally, we jointly train the entity pointer and MHS-based relation extractor via a combined loss with balancing ratio of 1:50.

\paragraph{Single Sentence Classification.} CHIP-CTC and KUAKE-QIC are tasks of this kind, which classifies a given sentence into one of a set of predefined categories. We simply build a softmax classifier on top of the final representation corresponding to the initial {\tt [CLS]} token for this classification task. 

\paragraph{Sentence Pair Classification.} The sentence pair matching tasks of CHIP-STS, KUAKE-QTR, and KUAKE-QQR, as well as the medical QA tasks of cMedQNLI and webMedQA are of this kind, aiming at predicting the semantic relationship between a pair of sentences according to a set of predefined labels. CHIP-CDN, after normalized terms have been retrieved for each original term, can also be formalized as a task of this kind, the aim of which is to judge if a normalized term matches the original term or not. Given a pair of sentences $(S_1, S_2)$, we pack them into a single input sequence ``{\tt[CLS]} $\!S_1\!$ {\tt[SEP]} $\!S_2\!$ {\tt[SEP]}'', and feed this sequence into a pre-trained encoder. Then we build a softmax classifier on top of {\tt[CLS]} representation to conduct sentence pair classification. For CHIP-CDN, we retrieve 100 candidate normalized terms for each original term from the whole ICD-10 vocabulary using Elasticsearch before pairwise classification.

\paragraph{Multiple Choice QA.} NLPEC is the only task of this kind. It selects the most plausible answer from 5 answer candidates for a given question. Textual evidences are also provided along with the question. Let $Q$ denote the question, $\{A_1, A_2, A_3, A_4, A_5\}$ the answer candidates, and $T$ the textual evidence. For each answer candidate $A_i$, we pack it with the question $Q$ and textual evidence $T$, and construct a single input sequence ``{\tt[CLS]} $\!A_i\!$ {\tt[SEP]} $\!Q\!$ {\tt[SEP]} $\!T\!$ {\tt[SEP]}''. We feed this sequence into a pre-trained encoder, and use {\tt[CLS]} representation to estimate if $A_i$ answers $Q$ given textual evidence $T$. In this fashion, we transform multiple choice into binary classification. At inference time, the candidate with highest probability is chosen as the correct answer.

\end{CJK}
\end{document}